\documentclass[runningheads]{llncs}

\usepackage{epsfig}
\usepackage{times}
\usepackage{graphicx}
\usepackage{amsmath}
\usepackage{amssymb}
\usepackage{bbm}
\usepackage{lib}
\usepackage{caption}
\usepackage{mathtools}
\usepackage{multirow}
\usepackage{epigraph}
\usepackage{booktabs}
\usepackage{microtype}
\usepackage{url}
\usepackage{color}
\usepackage[width=122mm,left=12mm,paperwidth=146mm,height=193mm,top=12mm,paperheight=217mm]{geometry}
\usepackage{wrapfig}

\usepackage[pagebackref=true,breaklinks=true,letterpaper=true,colorlinks,bookmarks=false]{hyperref}

\begin{document}
\newcommand\blfootnote[1]{%
  \begingroup
  \renewcommand\thefootnote{}\footnote{#1}%
  \addtocounter{footnote}{-1}%
  \endgroup
}

\title{Layer-structured 3D Scene Inference \\ via View Synthesis}

\titlerunning{Layer-structured 3D Scene Inference via View Synthesis}

\author{
Shubham Tulsiani$^{1*}$, Richard Tucker$^{2}$, Noah Snavely$^{2}$}

\institute{$^{1}$University of California, Berkeley ~~ $^{2}$Google\\
%{\email{shubhtuls@berkeley.edu, \{richardt,snavely\}@google.com}}
}

\authorrunning{S. Tulsiani, R. Tucker, N. Snavely }

\maketitle

\begin{abstract}
We present an approach to infer a layer-structured 3D representation of a scene from a single input image. This allows us to infer not only the depth of the visible pixels, but also to capture the texture and depth for content in the scene that is not directly visible. We overcome the challenge posed by the lack of direct supervision by instead leveraging a more naturally available multi-view supervisory signal. Our insight is to use view synthesis as a proxy task: we enforce that our representation (inferred from a single image), when  rendered from a novel perspective, matches the true observed image. We present a learning framework that operationalizes this insight using a new, differentiable novel view renderer. We provide qualitative and quantitative validation of our approach in two different settings, and demonstrate that we can learn to capture the hidden aspects of a scene. The project website can be found at \url{https://shubhtuls.github.io/lsi/}.

\end{abstract}

\section{Introduction}
\blfootnote{$^*$ The majority of the work was done while interning at Google.}
Humans have the ability to perceive beyond what they see, and to imagine the structure of the world even when it is not directly visible. Consider the image in \figref{teaser}. While we can clearly see a street scene with objects such as cars and trees, we can also reason about the shape and appearance of aspects of the scene hidden from view, such as the continuation of the buildings behind the trees, or the ground underneath the car.

While we humans can perceive the full 3D structure of a scene from a single image, scene representations commonly used in computer vision are often restricted to modeling the visible aspects, and can be characterized
% , borrowing Marr's terminology,
as \emph{2.5D representations}~\cite{Marr1982}. 2.5D representations such as depth maps are straightforward to use and learn because there is a one-to-one mapping between the pixels of an input image and the output representation. For the same reason, they also fail to allow for any extrapolation beyond what is immediately visible. In contrast, a robot or other agent might wish to predict the appearance of a scene from a different viewpoint, or reason about which parts of the scene are navigable. Such tasks are beyond what can be achieved in 2.5D.

In this work, we take a step towards reasoning about the 3D structure of scenes by learning to predict a \emph{layer-based} representation from a single image.  We use a representation known as a \emph{layered depth image} (LDI), originally developed in the computer graphics community~\cite{shade1998layered}. Unlike a depth map, which stores a single depth value per pixel, an LDI represents multiple ordered depths per pixel, along with an associated color for each depth, representing the multiple intersections of a ray with scene geometry (foreground objects, background behind those objects, etc.) In graphics, LDIs are an attractive representation for image-based rendering applications. For our purposes, they are also appealing as a 3D scene representation as they maintain the direct relationship between input pixels and output layers, while allowing for much more flexible and general modeling of scenes.

\begin{figure}[t]
\centering
\includegraphics[width=0.7\linewidth]{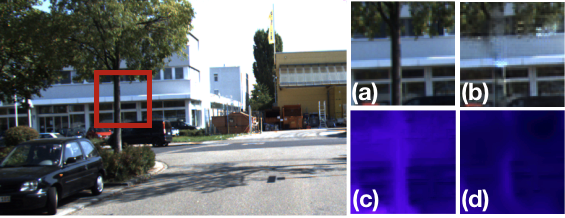}
\caption{\small \textbf{Perception beyond the visible.} On the left is an image of a street scene. While some parts of the scene are occluded, such as the building behind the tree highlighted by the red box, humans have no trouble reasoning about the shape and appearance of such hidden parts. In this work we go beyond 2.5D shape representations and learn to predict \emph{layered} scene representations from single images that capture more complete scenes, including hidden objects. On the right, we show our method's predicted 2-layer texture and shape for the highlighted area: a,b) show the predicted textures for the foreground and background layers respectively, and c,d) show the corresponding predicted inverse depth. Note how both predict structures behind the tree, such as the continuation of the building.}
\figlabel{teaser}
\end{figure}

A key challenge towards learning to predict such layered representations is the lack of available training data.  Our approach, depicted in \figref{overview}, builds on the insight that multiple images of the same scene, but from different views, can provide us with indirect supervision for learning about the underlying 3D structure. In particular, given two views of a scene, there will often be parts of the scene that are hidden from one view but visible from the second. We therefore use view synthesis as a proxy task: given a single input image, we predict an LDI representation and enforce that the novel views rendered using the prediction correspond to the observed reality.
%%%

In \secref{learning}, we present our learning setup that builds on this insight, and describe a training objective that enforces the desired prediction structure. To operationalize this learning procedure, we introduce an LDI rendering mechanism based on a new differentiable forward splatting layer. This layer may also be useful for other tasks at the intersection of graphics and learning. We then provide qualitative and quantitative validation of our approach in \secref{experiments} using two settings: a) analysis using synthetic data with known ground truth 3D, and b) a real outdoor driving dataset.
%%%%%%%%%%%%%%%%%%%%%%%%%%%%%%%%%%%%%%%%%%
%%%%%%%%%%%%%%%%%%%%%%%%%%%%%%%%%%%%%%%%%%
\begin{figure*}[t!]
\centering
\includegraphics[width=1.0\linewidth]{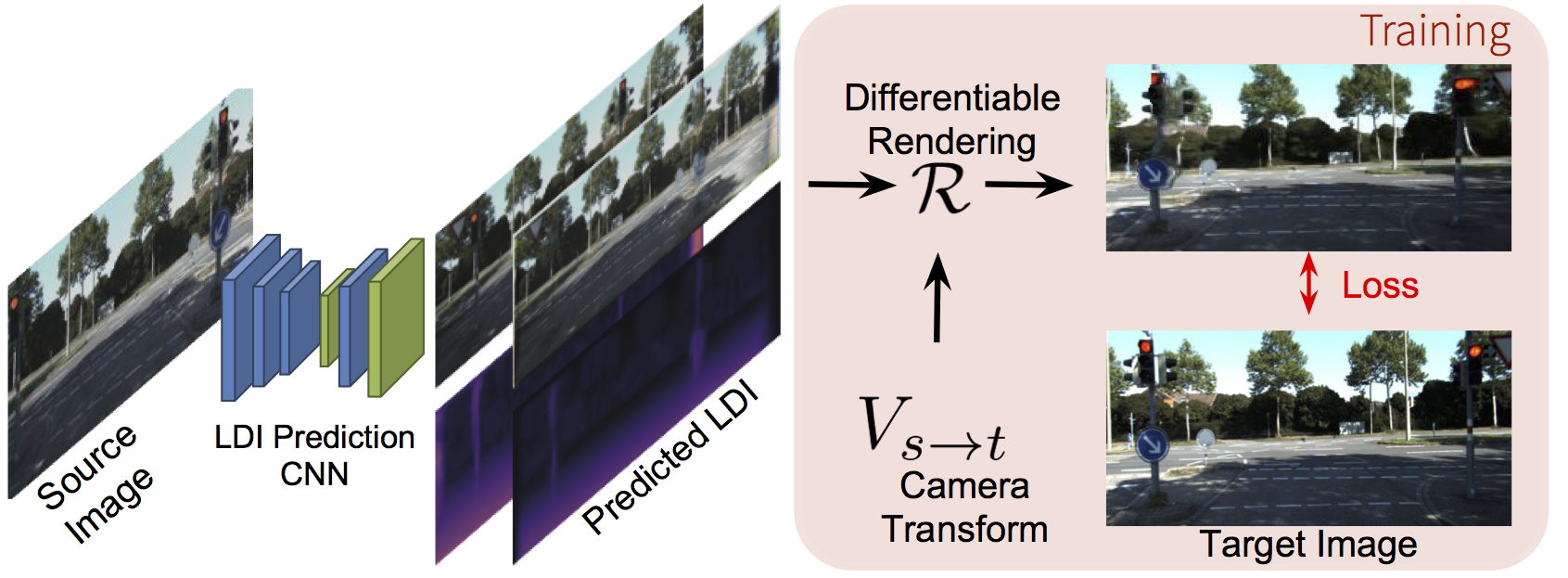}
\caption{\small \textbf{Approach overview.}
We learn a CNN that can predict, from a single input image, a layered representation of the scene (an LDI). During training, we leverage multi-view supervision using view synthesis as a proxy task, thereby allowing us to overcome the lack of direct supervision. While training our prediction CNN, we enforce that the predicted representation, when (differentiably) rendered from a novel view, matches the available target image.}
\figlabel{overview}
\end{figure*}
%%%%%%%%%%%%%%%%%%%%%%%%%%%%%%%%%%%%%%%%%%
%%%%%%%%%%%%%%%%%%%%%%%%%%%%%%%%%%%%%%%%%%

\section{Related Work}
\paragraph{\emph{\textbf{Single-view Depth/Surface Normal Prediction.}}}
Estimating pixel-wise depth and/or surface orientation has been a long-standing task in computer vision. Initial attempts treated geometric inference as a part of the inverse vision problem, leveraging primarily learning-free optimization methods for inference~\cite{sinha1993recovering,barron2015shape}. Over the years, the use of supervised learning has enabled more robust approaches~\cite{hoiem2005automatic,saxena2009make3d}, most recently with CNN-based methods~\cite{bansal2016marr,eigen2015predicting,wang2015designing}, yielding impressive results.

We also adopt a learning-based approach, but go beyond commonly used 2.5D representations that only infer shape for the \emph{visible} pixels. Some recent methods, with a similar goal, predict volumetric 3D from a depth image~\cite{song2016semantic}, or infer amodal aspects of a scene~\cite{Ehsani2017SeGANSA}. However, these methods require direct 3D supervision and are thus restricted to synthetically generated data. In contrast, our approach leverages indirect multi-view supervision that is more naturally obtainable, as well as ecologically plausible.

\paragraph{\emph{\textbf{Depth Prediction via View Synthesis.}}}
The challenge of leveraging indirect supervision for inference has been addressed by some recent multi-view supervised approaches. Garg \etal~\cite{garg2016unsupervised} and Godard \etal~\cite{godard2016unsupervised} used stereo images to learn a single-view depth prediction system by minimizing the inconsistency as measured by pixel-wise reprojection error. Subsequent works~\cite{vijayanarasimhan2017sfm,zhou2017unsupervised} further relax the constraint of having calibrated stereo images, and learn a single-view depth model from monocular videos.

We adopt a similar learning philosophy, \ie learning using multi-view supervision via view synthesis. However, our layered representation is different from the per-pixel depth predicted by these approaches, and in this work we address the related technical challenges. As we describe in \secref{learning}, our novel view rendering process is very different from the techniques used by these approaches.

\paragraph{\emph{\textbf{Multi-view Supervised 3D Object Reconstruction.}}} Learning-based approaches for single-view 3D object reconstruction have seen a similar shift in the forms of supervision required. Initial CNN-based methods~\cite{choy20163d,Girdhar16b} predicted voxel occupancy representations from a single input image but required full 3D supervision during training. Recent approaches have advocated alternate forms of supervision, \eg multi-view foreground masks ~\cite{rezende2016unsupervised,yan2016perspective,drcTulsiani17} or depth~\cite{drcTulsiani17}.

While these methods go beyond 2.5D predictions and infer full 3D structure, they use volumetric-occupancy-based representations that do not naturally extend to general scenes. The layered representations we use are instead closer to depth-based representations often used for scenes. Similarly, these methods commonly rely on cues like foreground masks from multiple views, which are more applicable to isolated objects than to complex scenes. In our scenario, we therefore rely only on multiple RGB images as supervision.

\paragraph{\emph{\textbf{Layered Scene Representations.}}}
Various layer-based scene representations are popular in the computer vision and graphics communities for reasons of parsimony, efficiency and descriptive power. Single-view based~\cite{hoiem2005automatic,isola2013scene,russell2009segmenting} or optical flow methods~\cite{wulff2017optical} often infer a parsimonious representation of the scene or flow by grouping the visible content into layers. While these methods do not reason about occlusion, Adelson~\cite{adelson1991layered} proposed using a planar layer-based representation to capture hidden surfaces and demonstrated that these can be inferred using motion~\cite{wang1993layered}. Similarly, Baker \etal~\cite{baker1998layered} proposed a stereo method that represents scenes as planar layers. Our work is most directly inspired by Shade \etal~\cite{shade1998layered}, who introduced the layered depth image (LDI) representation to capture the structure of general 3D scenes for use in image-based rendering.

We aim for a similar representation. However, in contrast to classical approaches that require multiple images for inference, we use machine learning to predict this representation from a single image at test time. Further, unlike previous single-view based methods, our predicted representation also reasons about occluded aspects of the scene.

\section{Learning LDI Prediction}
\seclabel{learning}
Our aim is to predict a 3D representation of a scene that includes not only the geometry of what we see, but also aspects of the scene not directly visible. A standard approach to geometric inference is to predict a depth map, which answers, for each pixel the question: \emph{`how far from the camera is the point imaged at this pixel?'}. In this work, we propose to predict a Layered Depth Image (LDI)~\cite{shade1998layered} representation that, in addition to the question above, also answers: \emph{`what lies behind the visible content at this pixel?'}.

As we do not have access to a dataset of paired examples of images with their corresponding LDI representations, we therefore exploit \emph{indirect} forms of supervision to learn LDI prediction. We note that since an LDI representation of a scene captures both visible and amodal aspects of a scene, it can allow us to geometrically synthesize novel views of the same scene, including aspects that are hidden to the input view. Our insight is that we can leverage \emph{view synthesis as a proxy target task}. We first formally describe our training setup and representation, then present our approach based on this insight. We also introduce a differentiable mechanism for rendering an LDI representation from novel views via a novel `soft z-buffering'-based forward splatting layer.

\subsection{Overview}
\paragraph{\emph{\textbf{Training Data.}}}
We leverage multi-view supervision to learn LDI prediction. Our training dataset is comprised of multiple scenes, with images from a few views available per scene. We assume a known camera transformation between the different images of the same scene. This form of supervision can easily be obtained using a calibrated camera rig, or by any natural agent which has access to its egomotion. Equivalently, we can consider the training data to consist of numerous source and target image pairs, where the two images in each pair are from the same scene and are related by a known transformation.

Concretely, we denote our training dataset of $N$ image pairs with associated cameras as $\{(I_s^n, I_t^n, \mathbf{K}_s^n, \mathbf{K}_t^n, \mathbf{R}^n, \mathbf{t}^n)\}_{n=1}^N$. Here $I_s^n, I_t^n$ represent two (source and target) images of the same scene, with camera intrinsics denoted as $\mathbf{K}_s^n, \mathbf{K}_t^n$ respectively. The relative camera transformation between the two image frames is captured by a rotation $\mathbf{R}^n$ and translation $\mathbf{t}^n$. We note that the training data leveraged does not assume any direct supervision for the scene's 3D structure.

\paragraph{\emph{\textbf{Predicted LDI Representation.}}}
A Layered Depth Image (LDI) representation (see \figref{ldi} for an illustration) represents the 3D structure of a scene using layers of depth and color images. An LDI representation with $L$ layers is of the form $\{(I^l, D^l)\}_{l=1}^L$. Here $(I^l, D^l)$ represent the texture (\ie, color) image $I$ and disparity (inverse depth) image $D$ corresponding to layer $l$. An important property of the LDI representation is that the structure captured in the layers is increasing in depth \ie for any pixel $p$, if $l_1 < l_2$, then $D^{l_1}(p) \ge D^{l_2}(p)$ (disparity is monotonically decreasing over layers, or, equivalently, depth is increasing). Therefore, the initial layer $l=1$ represents the visible content from the camera viewpoint (layers in an LDI do not have an alpha channel or mask). In fact, a standard depth map representation can be considered as an LDI with a single layer, with $I^1$ being the observed image.

%%%%%%%%%%%%%%%%%%%%%%%%%%%%%%%%%%%%%%%%%%
%%%%%%%%%%%%%%%%%%%%%%%%%%%%%%%%%%%%%%%%%%
\begin{figure}[h!]
\centering
\includegraphics[width=0.6\linewidth]{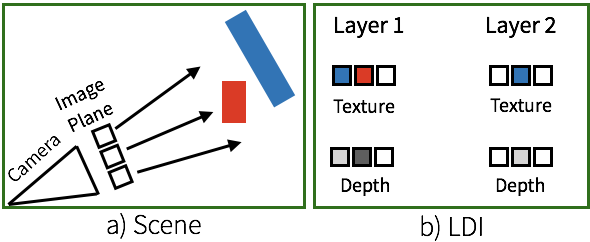}
\caption{\small \textbf{Layered Depth Images (LDIs).}
Illustration of a layered depth image (LDI) for a simple scene. The first layer captures the depth (darker indicates closer) and texture of the visible points, and the second layer describes the occluded structure.}
\figlabel{ldi}
\end{figure}

%\begin{wrapfigure}{r}{0.4\columnwidth}
%\vspace{-4mm}
%\includegraphics[width=1\linewidth]{figures/ldi_illustration.png}
%\caption{\small \textbf{Layered Depth Images (LDIs).}
%Illustration of a layered depth image (LDI) for a simple scene. The first layer captures the depth (darker indicates closer) and texture of the directly visible points, and the second layer describes the occluded structure.}
%\figlabel{ldi}
%\end{wrapfigure}
%%%%%%%%%%%%%%%%%%%%%%%%%%%%%%%%%%%%%%%%%%
%%%%%%%%%%%%%%%%%%%%%%%%%%%%%%%%%%%%%%%%%%

In our work, we aim to learn an LDI prediction function $f$, parametrized as a CNN $f_{\theta}$, which, given a single input image $I$, can infer the corresponding LDI representation $\{(I^l, D^l)\}_{l=1}^L$. Intuitively, the first layer corresponds to the aspects of the scene visible from the camera viewpoint, and the subsequent layers capture aspects occluded in the current view. Although in this work we restrict ourselves to inferring two layers, the learning procedure presented is equally applicable for the more general scenario.

\paragraph{\emph{\textbf{View Synthesis as Supervision.}}}
Given a source image $I_s$, we predict the corresponding LDI representation $f_{\theta}(I_s) = \{(I_s^l, D_s^l)\}_{l=1}^L$. During training, we also have access to an image $I_t$ of the same scene as $I_s$, but from a different viewpoint. We write $V_{s \rightarrow t} \equiv (\mathbf{K}_s, \mathbf{K}_t, \mathbf{R}, \mathbf{t})$ to denote the camera transform between the source frame and the target frame, including intrinsic and extrinsic camera parameters. With this transform and our predicted LDI representation, we can render a predicted image from the target viewpoint. In particular, using a geometrically defined rendering function $\mathcal{R}$, we can express the novel target view rendered from the source image as $\mathcal{R}(f_{\theta}(I_s); V_{s \rightarrow t})$.

We can thus obtain a learning signal for our LDI predictor $f_{\theta}$ by enforcing similarity between the predicted target view $\mathcal{R}(f_{\theta}(I_s); V_{s \rightarrow t})$ and the observed target image $I_t$. There are two aspects of this learning setup that allow us to learn meaningful prediction: a) the novel view $I_t$ may contain new scene content compared to $I_s$, \eg dis\-occluded regions, therefore the LDI $f_{\theta}(I_s)$ must capture more than the visible structure; and b) the LDI $f_{\theta}(I_s)$ is predicted \emph{independently} of the target view/image $I_t$ which may be sampled arbitrarily, and hence the predicted LDI should be able to explain content from many possible novel views.
%Further, as the rendering process $\mathcal{R}$ is geometrically defined and implicitly treats the predictions $f_{\theta}$ as layer-wise depths and textures, the CNN $f_{\theta}$ learns to predict outputs consistent with this interpretation even though this is not enforced via any direct form of supervision.

\paragraph{\emph{\textbf{The need for forward-rendering.}}}  As noted by Shade \etal when introducing the LDI representation~\cite{shade1998layered}, the rendering process for synthesizing a novel view given a source LDI requires forward-splatting-based rendering. This requirement leads to a subtle but important difference in our training procedure compared to prior multi-view supervised depth prediction methods~\cite{garg2016unsupervised,godard2016unsupervised,zhou2017unsupervised}: while prior approaches rely on inverse warping for rendering, our representation necessitates the use of forward rendering.

Concretely, prior approaches, given a source image $I_s$, predict a per-pixel depth map. Then, given a novel view image, $I_t$, they reconstruct the source image by `looking up' pixels from $I_t$ via the predicted depth and camera transform. Therefore, the `rendered view' is the same as the input view for which the geometry is inferred, \ie these methods do not render a novel view, but instead re-render the source view. This procedure only enforces that correct geometry is learned for pixels visible to both views.

However, in our scenario, since we explicitly want to predict beyond the visible structure, we cannot adopt this approach. Instead, we synthesize novel views using our layered representation, thereby allowing us to learn about both the visible and the occluded scene structure. This necessitates forward rendering, \ie constructing a target view given the source view texture and geometry, as opposed to inverse warping, \ie reconstructing a source view by using source geometry and target frame texture.

\subsection{Differentiable Rendering of an LDI}
\seclabel{viewsynth}
Given a predicted LDI representation $\{(I^l_s, D^l_s)\}$ in a source image frame, we want to render a novel viewpoint related by a transform $V_{s \rightarrow t}$. We do so by treating the LDI as a textured point cloud, with each pixel in each layer corresponding to a point. We first forward-project each source point onto the target frame, then handle occlusions by proposing a `soft z-buffer', and finally render the target image by a weighted average of the colors of projected points.

\paragraph{\emph{\textbf{Forward Projection.}}}
Denoting by $p_s^l$ the pixel $p_s \equiv (x_s, y_s)$ in layer $l$, we can compute its projected position and inverse depth in the target frame coordinates using the (predicted) inverse depth $d_s^l \equiv D^l_s(p_s)$ and the camera parameters.
\begin{gather}
    \begin{bmatrix}
    \bar{x}_t(p_s^l) \\ 
    \bar{y}_t(p_s^l) \\
    1 \\
    \bar{d}_t(p_s^l)
    \end{bmatrix} \sim
    \begin{bmatrix}
    \mathbf{K}_t & \hat{0} \\ 
    \hat{0} & 1
    \end{bmatrix}~
    \begin{bmatrix}
    \mathbf{R} & \hat{\mathbf{t}} \\ 
    \hat{0} & 1
    \end{bmatrix}
    ~
    \begin{bmatrix}
    \mathbf{K}_s^{-1} & \hat{0} \\ 
    \hat{0} & 1
    \end{bmatrix}~
    \begin{bmatrix}
    x_s \\ 
    y_s \\
    1 \\
    d_s^l
    \end{bmatrix}
\end{gather}

\paragraph{\emph{\textbf{Splatting with soft z-buffering.}}}
Using the above transformation, we can \emph{forward splat} this point cloud to the target frame. Intuitively, we consider the target frame image as an empty canvas. Then, each source point $p_s^l$ adds paint onto the canvas, but only at the pixels immediately around its projection. Via this process, many source points may contribute to the same target image pixel, and we want the closer ones to occlude the further ones. In traditional rendering, this can be achieved using a z-buffer, with only the closest point contributing to the rendering of a pixel.

However, this process results in a discontinuous and non-differentiable rendering function that is unsuitable for our framework. Instead, we propose a \emph{soft} z-buffer using a weight $w(p_t, p_s^l)$ that specifies the contribution of $p_s^l$ to the target image pixel $p_t$. Defining $\mathcal{B}(x_0, x_1)\equiv \max~(0, 1 - |x_0 - x_1|)$, we compute the weights as:
\begin{gather}
w(p_t, p_s^l) = \exp\left(\frac{\bar{d}_t(p_s^l)}{\tau}\right)~\mathcal{B}(\bar{x}_t(p_s^l), x_t)~\mathcal{B}(\bar{y}_t(p_s^l), y_t)
\end{gather}
The initial exponential factor, modulated by the temperature $\tau$, enforces higher precedence for points closer to the camera. A large value of  $\tau$ results in `softer' z-buffering, whereas a small value yields a rendering process analogous to standard z-buffering. The latter terms simply represent bilinear interpolation weights and ensure that each source point only contributes non-zero weight to target pixels in the immediate neighborhood.

\paragraph{\emph{\textbf{Rendering.}}}
Finally, we compute the rendered texture $\bar{I}_t(p_t)$ at each target pixel $p_t$ as a weighted average of the contributions of points that splat to that pixel:
\begin{gather}
\eqlabel{itpt}
\bar{I}_t(p_t) = \frac{\sum_{p_s^l} I_s^l~w(p_t, p_s^l)  ~~ + ~~ \epsilon}{\sum_{p_s^l} w(p_t, p_s^l) ~~ + ~~ \epsilon}
\end{gather}
The small $\epsilon$ in the denominator ensures numerical stability for target pixels that correspond to no source point. A similar term in the numerator biases the color for such pixels towards white. All operations involved in rendering the novel target view are differentiable, including the forward projection, depth-dependent weight computation, and final color computation. Hence, we can use this \emph{rendering via forward splatting} process as the differentiable $\mathcal{R}(f_{\theta}(I_s); V_{s \rightarrow t})$ required in our learning framework.

\subsection{Network Architecture}
We adopt the DispNet~\cite{mayer2016large} architecture for our LDI prediction CNN shown in \figref{architecture}. Given the input color image, a convolutional encoder processes it to compute spatial features at various resolutions. We then decode these via upconvolutions to get back to the image resolution. Each layer in the decoder also receives the features from the corresponding encoder layer via skip connections. While we use a single CNN to predict disparities and textures for all LDI layers, we find it critical to have disjoint prediction branches to infer each LDI layer. We hypothesize that this occurs because the foreground layer gets more learning signal, and sharing all the prediction weights makes it difficult for the learning signals for the background layer to compete. Therefore, the last three decoding blocks and final prediction blocks are independent for each LDI layer.

\subsection{Training Objective}
To train our CNN $f_{\theta}$, we use view synthesis as a proxy task: given a source image $I_s$, we predict a corresponding LDI and render it from a novel viewpoint. As a training objective, we enforce that this rendered image should be similar to the observed image from that viewpoint. However, there are some additional nuances that we need to consider when formulating our training objective.

\paragraph{\emph{\textbf{Depth Monotonicity.}}}
The layers in our LDI representation are supposed to capture content at increasing depths. We therefore enforce that the inverse depth across layers at any pixel is non-increasing:
\begin{gather}
    L_{\mathrm{inc}}(I_s) = \sum_{p_s, l} \max(0, D_s^{l+1}(p_s) - D_s^l(p_s)).
\end{gather}

\paragraph{\emph{\textbf{Consistency with Source.}}}
The typical LDI representation enforces that the first layer's texture corresponds to the observed source. We additionally enforce a similar constraint even for background layers when the predicted geometry is close to the foreground layer.
We compute a normalized weight for the layers at each pixel, denoted as $w(p_s, l) \propto \exp{\frac{D_s^l(p_s)}{\tau}}$, and define a weighted penalty for deviation from the observed image:
\begin{gather}
    L_{\mathrm{sc}}(I_s) = \sum_{p_s, l} w(p_s, l) \|I_s(p_s) - I_s^l(p_s) \|_1.
\end{gather}
This loss encourages the predicted texture at each layer to match the source texture, while allowing significant deviations in case of occlusions, \ie where the background layer is much further than the foreground. In conjunction with $L_{\mathrm{inc}}$, this loss enforces that the predicted representation adheres to the constraints of being an LDI.

%%%%%%%%%%%%%%%%%%%%%%%%%%%%%%%%%%%%%%%%%%
%%%%%%%%%%%%%%%%%%%%%%%%%%%%%%%%%%%%%%%%%%
\begin{figure}[t!]
\centering
\includegraphics[width=0.7\linewidth]{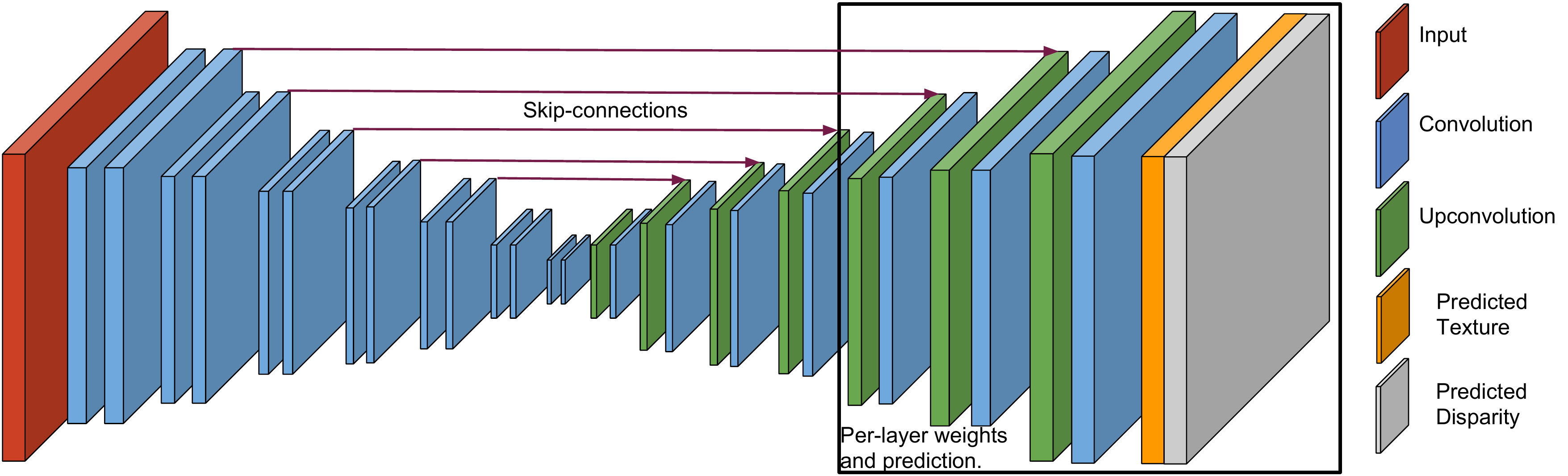}
\caption{\small \textbf{Overview of our CNN architecture.}
We take as input an image and predict per-layer texture and inverse depth. Our CNN architecture consists of a convolutional encoder and decoder with skip-connections. We use disjoint prediction branches for inferring the texture and depth for each LDI layer.}
\figlabel{architecture}
\end{figure}
%%%%%%%%%%%%%%%%%%%%%%%%%%%%%%%%%%%%%%%%%%
%%%%%%%%%%%%%%%%%%%%%%%%%%%%%%%%%%%%%%%%%%

\paragraph{\emph{\textbf{Allowing Content Magnification.}}}
The forward-splatting rendering method described in \secref{viewsynth} computes a novel view image by splatting each source pixel onto the target frame. This may result in `cracks'~\cite{grossman1998point}---target pixels that are empty because no source pixels splat onto them. For example, if the target image contains a close-up view of an object that is faraway in the source image, too few source points will splat into that large target region to cover it completely. To overcome this, we simply render the target frame at half the input resolution, \ie the output image from the rendering function described in \secref{viewsynth} is half the size of the input LDI.

\paragraph{\emph{\textbf{Ignoring Image Boundaries.}}} While an LDI representation can explain the dis\-occluded content that becomes visible in a novel view, it cannot capture the pixels in the target frame that are outside the image boundary in the source frame. We would like to ignore such pixels in the view synthesis loss. However, we do not have ground-truth to tell us which pixels these are. Instead, we use the heuristic of ignoring pixels around the boundary. Denoting as $M$ a binary mask that is zero around the image edges, we define our view synthesis loss as:
\begin{multline}
\eqlabel{comp_sp_loss}
    L_{\mathrm{vs}}(I_s, I_t, V_{s \rightarrow t}) = \| M \odot I_t - M \odot \bar{I}_t \|_1 ~~~~ \text{where} ~~ \bar{I}_t = \mathcal{R}(f_{\theta}(I_s); V_{s \rightarrow t}).
\end{multline}
As described above, the rendered image $\bar{I}_t$ and the target image $I_t$ are spatially smaller than $I_s$.

\paragraph{\emph{\textbf{Overcoming Depth Precedence.}}}
Consider synthesizing pixel $p_t$ as described in \eqref{itpt}. While the weighted averaging across layers resembles z-buffer-based rendering, it has the disadvantage of making it harder to learn a layer if there is another preceding (and possibly incorrectly predicted) layer in front of it. To overcome this, and therefore to speed up the learning of layers independent of other layers, we add an additional loss term. Denoting as $\bar{I}_t^l$ a target image rendered using \emph{only layer $l$}, we add an additional `min-view synthesis' loss measuring the minimum pixel-wise error across per-layer synthesized views:
\begin{gather}
    L_{\mathrm{m-vs}}(I_s, I_t, V_{s \rightarrow t}) = \sum_{p_t} \min_l M(p_t)  \|I_t(p_t) - \bar{I}^l_t(p_t) \|_1
\end{gather}
In contrast to the loss in \eqref{comp_sp_loss}, which combines the effects of all layers when measuring the reconstruction error at $p_t$, this loss term simply enforces that at least one layer should correctly explain the observed $I_t(p_t)$. Therefore, a background layer can still get a meaningful learning signal even if there is a foreground layer incorrectly occluding it. Empirically, we found that this term is crucial to allow for learning the background layer.

\paragraph{\emph{\textbf{Smoothness.}}} We use a depth smoothness prior $L_{\mathrm{sm}}$ which minimizes the $L_1$ norm of the second-order spatial derivatives of the predicted inverse depths $D_s^l$.

Our final learning objective, combining the various loss terms defined above (with different weights) is:
\begin{gather}
    L_{\mathrm{final}} = L_{\mathrm{vs}} + L_{\mathrm{m-vs}} + L_{\mathrm{sc}} + L_{\mathrm{inc}} + L_{\mathrm{sm}}
\end{gather}
Using this learning objective, we can train our LDI prediction CNN $f_{\theta}$ using a dataset comprised only of paired source and target images of the same scene.
%%%%%%%%%%%%%%%%%%%%%%%%%%%%%%%%%%%
%%%%%%%%%%%%%%%%%%%%%%%%%%%%%%%%%%%

\section{Experiments}
\seclabel{experiments}
We consider two different scenarios to learn single-view inference of a layer-structured scene representation. We first study our approach in a synthetic, but nevertheless challenging, setting using procedurally generated data. We then use our method to learn from stereo pairs in an outdoor setting.

\subsection{Analysis using Synthetic Data}
In order to examine our method in a controlled setting with full knowledge of the underlying 3D scene structure, we create a dataset of procedurally generated scenes. We first describe the details of the generation process, and then discuss the training details and our results.

%%%%%%%%%%%%%%%%%%%%%%%%%%%%%%%%%%%%%%%%%%
%%%%%%%%%%%%%%%%%%%%%%%%%%%%%%%%%%%%%%%%%%
\begin{figure}[t]
\centering
\includegraphics[width=0.9\linewidth]{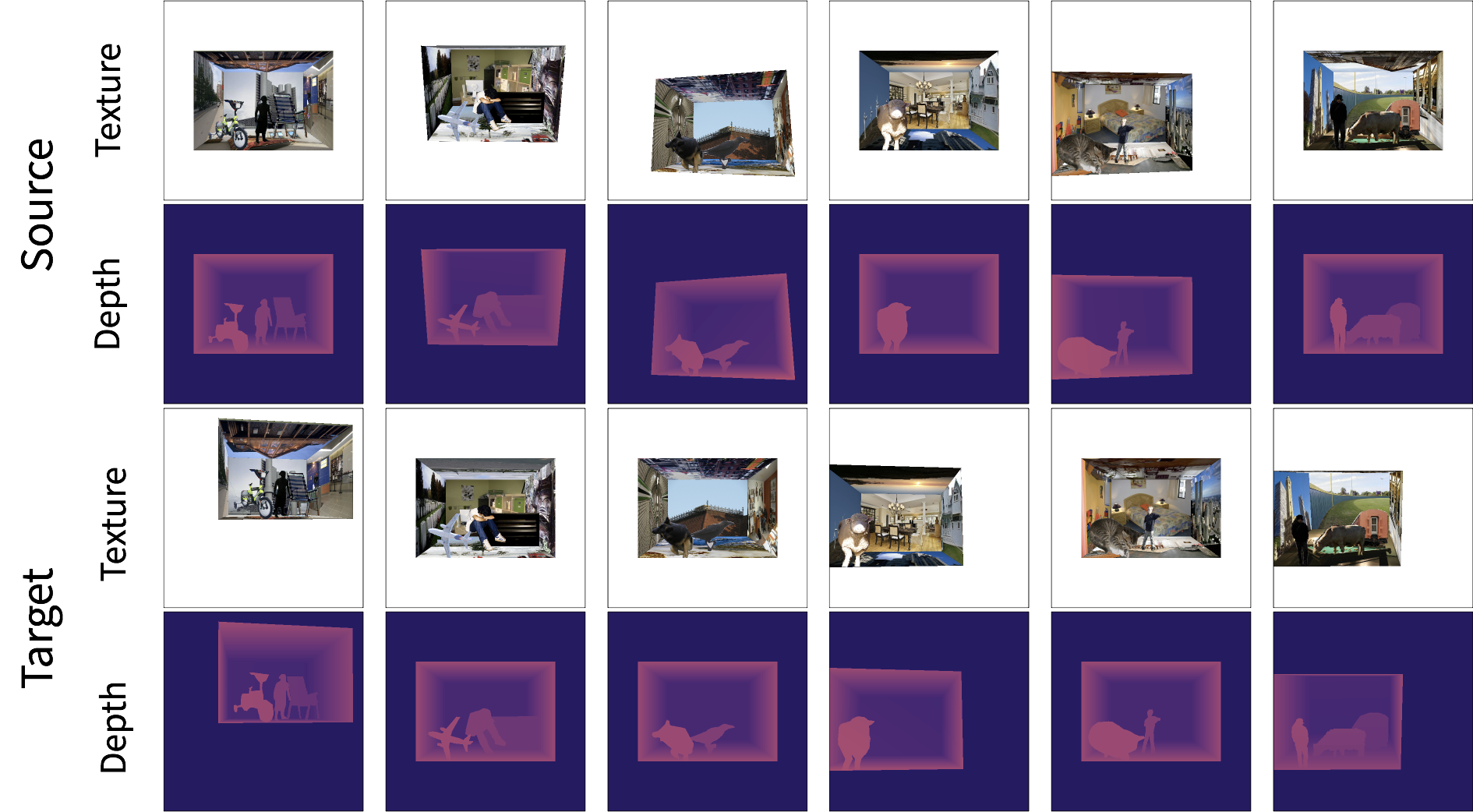}
\caption{\small \textbf{Procedurally generated synthetic data.}
We show 6 random training samples (top: source image and corresponding inverse depth, bottom: target image with corresponding inverse depth). Note that only the color images are used for learning.}
\figlabel{synthsamples}
\end{figure}
%%%%%%%%%%%%%%%%%%%%%%%%%%%%%%%%%%%%%%%%%%
%%%%%%%%%%%%%%%%%%%%%%%%%%%%%%%%%%%%%%%%%%

\paragraph{\emph{\textbf{Dataset.}}}
We generate our synthetic data to have a room-like layout with two side `walls', one back `wall', a `ceiling' and a `floor'. We additionally place one to three upright segmented objects on the floor. The `room' box is always at a fixed location in the world frame, and is of a fixed size. The segmented foreground objects are randomly placed, from left to right, at increasing depths and lie on a front-facing planar surface. To obtain the foreground objects, we randomly sample from the unoccluded and untruncated object instances in the PASCAL VOC dataset~\cite{pascalvoc2012}. The textures on the room walls are obtained using random images from the SUN 2012 dataset~\cite{xiao2010sun}.

%%%%%%%%%%%%%%%%%%%%%%%%%%%%%%%%%%%%%%%%%%
%%%%%%%%%%%%%%%%%%%%%%%%%%%%%%%%%%%%%%%%%%
\begin{figure}[t]
\centering
\includegraphics[width=1\textwidth]{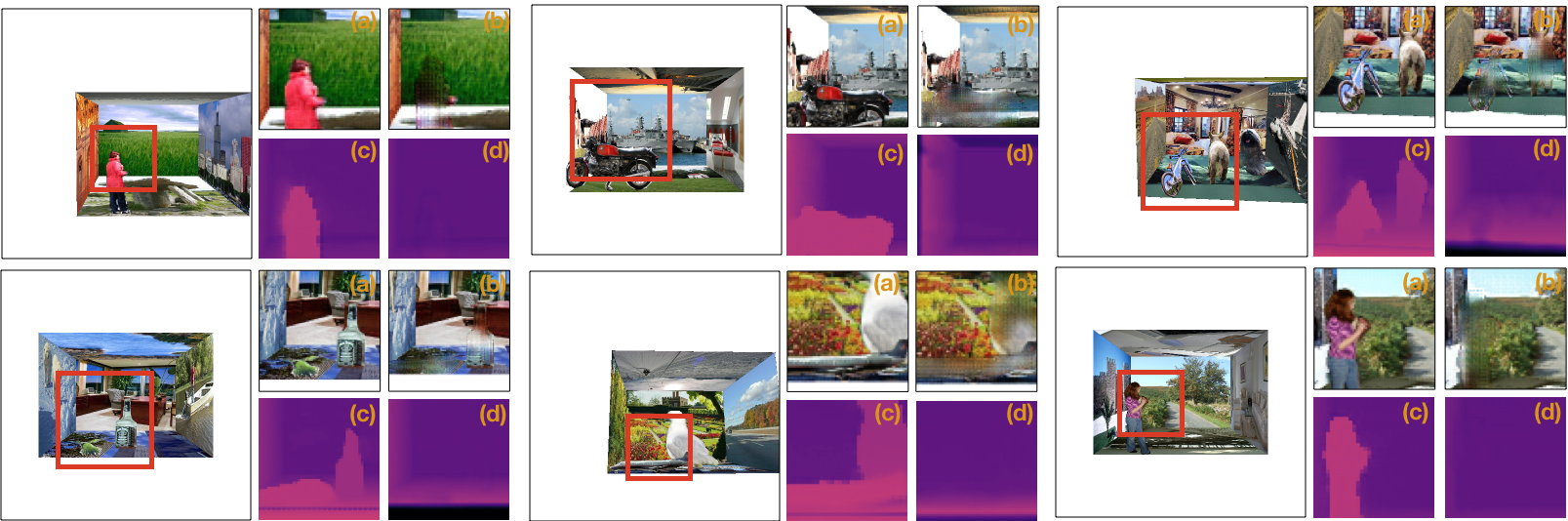}
\caption{\small \textbf{Sample LDI prediction results on synthetic data.}
For each input image on the left, we show our method's predicted 2-layer texture and geometry for the highlighted area: a,b) show the predicted textures for the foreground and background layers respectively, and c,d) depict the corresponding predicted disparity.}
\figlabel{synth_results}
\end{figure}
%%%%%%%%%%%%%%%%%%%%%%%%%%%%%%%%%%%%%%%%%%
%%%%%%%%%%%%%%%%%%%%%%%%%%%%%%%%%%%%%%%%%%
To sample the source and target views for training our LDI prediction, we randomly assign one of them to correspond to the canonical front-facing world view. The other view corresponds to a random camera translation with a random rotation. We ensure that the transformation can be large enough such that the novel views can often image the content behind the foreground object(s) in the source view. We show some sample source and target pairs in \figref{synthsamples}.

Note that while the geometry of the scene layout is relatively simple, the foreground objects can have differing shapes due their respective segmentation. Further, the surface textures are drawn from diverse real images and significantly add to the complexity, particularly as our aim is to infer both the geometry and the texture for the scene layers.

\paragraph{\emph{\textbf{Training Details.}}}
We split the PASCAL VOC objects and the SUN 2012 images into random subsets corresponding to a train/validation/test split of $70\% - 15\% - 15\%$. We use the corresponding images and objects to generate training samples to train our LDI prediction CNN $f_{\theta}$. We train our CNN for 600k iterations using the ADAM optimizer~\cite{kingma2014adam}. Based on the dataset statistics, we restrict the maximum inverse depth predicted to correspond to 1m.

%%%%%%%%%%%%%%%%%%%%%%%%%%%%%%%%%%%%%%%%%%
%%%%%%%%%%%%%%%%%%%%%%%%%%%%%%%%%%%%%%%%%%
\begin{figure}[t]
\centering
\includegraphics[width=1\textwidth]{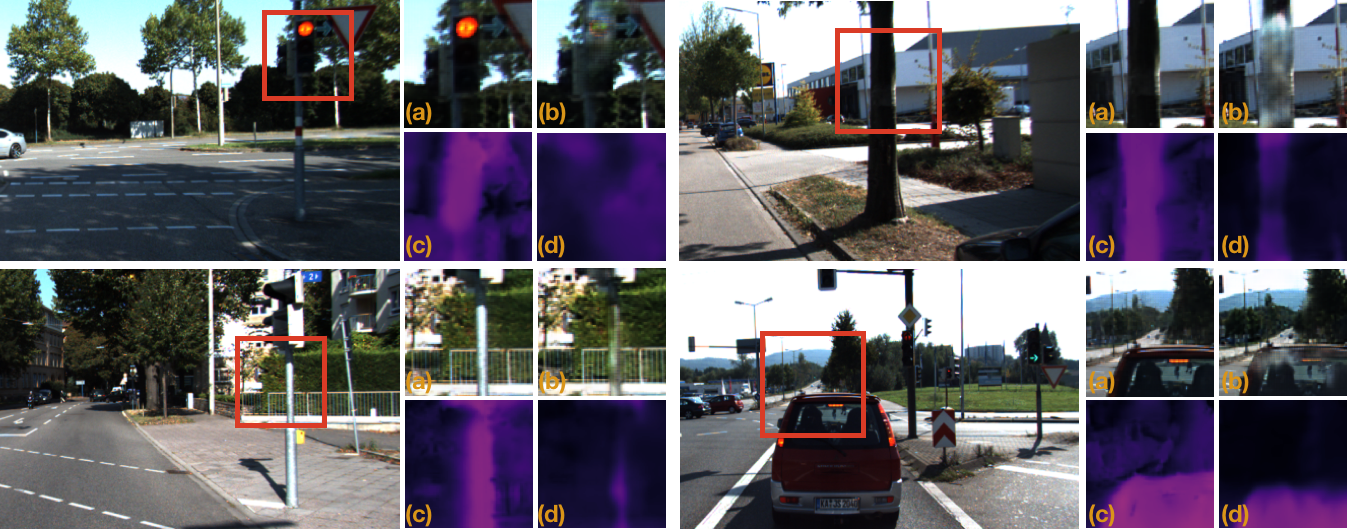}
\caption{\small \textbf{Sample LDI prediction results on the KITTI dataset.}
For each input image on the left, we show our method's predicted 2-layer texture and geometry for the highlighted area: a, b) show the predicted textures for the foreground and background layers respectively, and c,d) depict the corresponding predicted disparity.}
\figlabel{kitti_results}
\end{figure}
%%%%%%%%%%%%%%%%%%%%%%%%%%%%%%%%%%%%%%%%%%
%%%%%%%%%%%%%%%%%%%%%%%%%%%%%%%%%%%%%%%%%%

\paragraph{\emph{\textbf{Results.}}} We visualize the predictions of our learned LDI prediction CNN in \figref{synth_results}. We observe that it is able to predict the correct geometry for the foreground layer \ie per-pixel depth. More interestingly, it can leverage the background layer to successfully infer the geometry of the occluded scene content and hallucinate plausible corresponding textures. We observe some interesting error modes in the prediction, \eg incorrect background layer predictions at the base of wide objects, or spurious details in the background layer at pixels outside the `room'. Both these occur because we do not use any direct supervision for learning, but instead rely on a view synthesis loss. The first error mode occurs because we never fully `see behind' the base of wide objects even in novel views. Similarly, the spurious details are only present in regions which are consistently occluded by the foreground layer and therefore ignored for view synthesis.

%%%%%%%%%%%%%%%%%%%%%%%%%%%%%%%%%%%%%%%%%%
%%%%%%%%%%%%%%%%%%%%%%%%%%%%%%%%%%%%%%%%%%
\renewcommand{\arraystretch}{1.2}
\setlength{\tabcolsep}{4pt}

\begin{table}[t]
\centering
\begin{tabular}{l c c c}
\toprule
View Synthesis Error & All Pixels & Dis-occluded Pixels \\
\midrule
1 layer model  & 0.0398  & 0.1439  \\
2 layer model & 0.0392  & 0.1301 \\
\bottomrule
\end{tabular}
\vspace{2mm}
\caption{\small \textbf{View synthesis error on synthetic data.} We compare our 2 layer LDI prediction CNN  against a single layer model that can only capture the visible aspects. We report the mean pixel-wise $\ell_1$ error between the ground-truth novel view and the corresponding view rendered using the predicted representations.}
\tablelabel{synthviewsynth}
\vspace{-3mm}
\end{table}
%%%%%%%%%%%%%%%%%%%%%%%%%%%%%%%%%%%%%%%%%%
%%%%%%%%%%%%%%%%%%%%%%%%%%%%%%%%%%%%%%%%%%

%%%%%%%%%%%%%%%%%%%%%%%%%%%%%%%%%%%%%%%%%%
%%%%%%%%%%%%%%%%%%%%%%%%%%%%%%%%%%%%%%%%%%
\renewcommand{\arraystretch}{1.2}
\setlength{\tabcolsep}{4pt}

\begin{table}[t]
\centering
\begin{tabular}{l c c c}
\toprule
Inverse Depth & Foreground Layer & Background Layer \\
Error & (All Pixels) & (Hidden Pixels) \\
\midrule
1 layer model  & 0.0092  & 0.1307 (*) \\
2 layer model & 0.0102  & 0.0152 \\
\bottomrule
\end{tabular}
\vspace{2mm}
\caption{\small \textbf{Geometry prediction error on synthetic data.} We measure mean pixel-wise error in the predicted inverse depth(s) against the ground-truth. (*) As the single layer model does not infer background, we evaluate its error for the background layer using the foreground depth predictions. This serves to provide an instructive upper bound for the error of the LDI model.}
\tablelabel{synthdepth}
\vspace{-3mm}
\end{table}
%%%%%%%%%%%%%%%%%%%%%%%%%%%%%%%%%%%%%%%%%%
%%%%%%%%%%%%%%%%%%%%%%%%%%%%%%%%%%%%%%%%%%

We analyze our learned representation by evaluating how well we can synthesize novel views using it. We report in \tableref{synthviewsynth} the mean $\ell_1$ error for view synthesis and compare our 2 layer model vs a single layer model also trained for the view synthesis task, using the same architecture and hyper-parameters. Note that that single layer model can only hope to capture the visible aspects, but not the occluded structure.  We observe that we perform slightly better than the single layer model. Since most of the scene pixels are visible in both, the source and target views, a single layer model explains them well. However, we see that the error difference is more significant if we restrict our analysis to only the dis-occluded pixels \ie pixels in the target image which are not visible in the source view. This supports the claim that our predicted LDI representation does indeed capture more than the directly visible structure.

We also report in \tableref{synthdepth} the error in the predicted inverse depth(s) against the known ground-truth. We restrict the error computation for the background layer to pixels where the depth differs from the foreground layer. Since the one layer model only captures the foreground, and does not predict the background depths, we measure its error for the background layer using the foreground layer predictions. While this is an obviously harsh comparison, as the one layer model, by design, cannot capture the hidden depth,  the fact that our predicted background layer is `closer' serves to empirically show that our learned model infers  meaningful geometry for the background layer.

\subsection{Experiments on KITTI}
We demonstrate the applicability of our framework in a more realistic setting: outdoor scenes with images collected using a calibrated stereo camera setup. We note that previous methods applied to this setting have been restricted to inferring the depth of the visible pixels, and that it is encouraging that we can go beyond this representation.

\paragraph{\emph{\textbf{Dataset.}}} We use the `raw' sequences from the KITTI dataset~\cite{Geiger2013IJRR}, restricting our data to the 30 sequences from the \emph{city} category as these more often contain interesting occluders \eg people or traffic lights. The multi-view supervision we use corresponds to images from calibrated stereo cameras that are 0.5m apart. We use both the left and the right camera images as source images, and treat the other as the target view for which the view synthesis loss is minimized. Due to the camera setup, the view sampling corresponds to a lateral motion of 0.5m and is more restrictive compared to the synthetic data.

\paragraph{\emph{\textbf{Training Details.}}}
We randomly choose 22 among the 30 \emph{city} sequences for training, and use 4 each for validation and testing. This results in a training set of about 6,000 stereo pairs. We use similar hyper-parameters and optimization algorithm to the synthetic data scenario, but alter the closest possible depth to correspond to 2m.

%%%%%%%%%%%%%%%%%%%%%%%%%%%%%%%%%%%%%%%%%%
%%%%%%%%%%%%%%%%%%%%%%%%%%%%%%%%%%%%%%%%%%
\renewcommand{\arraystretch}{1.2}
\setlength{\tabcolsep}{4pt}

\begin{table}[t]
\centering
\begin{tabular}{l c c c}
\toprule
View Synthesis Error & All Pixels & Dis-occluded Pixels \\
\midrule
1 layer model  & 0.0583 & 0.0813  \\
2 layer model & 0.0581 & 0.0800  \\
\bottomrule
\end{tabular}
\vspace{2mm}
\caption{\small \textbf{View synthesis error on KITTI.} We compare our 2 layer LDI prediction CNN  against a single layer model that can only capture the visible aspects. We report the mean pixel-wise view synthesis error when rendering novel views using the predicted representations.}
\tablelabel{kitti_results}
\vspace{-3mm}
\end{table}
%%%%%%%%%%%%%%%%%%%%%%%%%%%%%%%%%%%%%%%%%%
%%%%%%%%%%%%%%%%%%%%%%%%%%%%%%%%%%%%%%%%%%

\paragraph{\emph{\textbf{Results.}}}
We visualize sample predictions of our learned LDI prediction CNN in \figref{kitti_results}. We observe that it is able to predict the correct geometry for the foreground layer \ie per-pixel depth. Similar to the synthetic data scenario, we observe that it can leverage the background layer to hallucinate plausible geometry and texture of the occluded scene content, although to a lesser extent. We hypothesize that the reduction in usage of the background layer is because the view transformation between the source and target views is small compared to the scene scale, and we therefore only infer background layer mostly corresponding to a) thin scene structures smaller than the stereo baseline, or b) around the boundaries of larger objects/structures \eg cars.

We do not have the full 3D structure of the scenes to compare our predicted LDI against, but we can evaluate the ability of this representation to infer the available novel views, and we report these evaluations in \tableref{kitti_results}. As we do not have the ground-truth for the dis-occluded pixels, we instead use the unmatched pixels from an off-the-shelf stereo matching algorithm~\cite{Yamaguchi2014EfficientJS}. This algorithm, in addition to computing disparity, attempts to identify pixels with no correspondence in the other view, thus providing (approximate) dis-occlusion labels (see supplementary material for visualizations). Measuring the pixel-wise reconstruction error, we again observe that our two-layer LDI model performs slightly better than a single layer model which only models the foreground. Additionally, the difference is a bit more prominent for the dis-occluded pixels.

While the above evaluation indicates our ability to capture occluded structure, it is also worth examining the accuracy of the predicted depth. To this end, we compared results on our test set against the publicly available model from Zhou \etal~\cite{zhou2017unsupervised}, since we use a similar CNN architecture facilitating a more apples-to-apples comparison. We perform comparably, achieving an Absolute Relative error of 0.1856, compared to an error of 0.2079 by~\cite{zhou2017unsupervised}. While other monocular depth estimation approaches can further achieve improved results using stronger supervision, better architectures or cycle consistency~\cite{godard2016unsupervised}, we note that achieving state-of-the-art depth prediction is not our central goal. However, we find it encouraging that our proposed LDI prediction approach does yield somewhat competitive depth prediction results.

\section{Discussion}
We have presented a learning-based method to infer a layer-structured representation of scenes that can go beyond common 2.5D representations and allow for reasoning about occluded structures. There are, however, a number of challenges yet to be addressed. As we only rely on multi-view supervision, the learned geometry is restricted by the extent of available motion across training views. Additionally, it would be interesting to extend our layered representation to include a notion of grouping, incorporate semantics and semantic priors (\eg `roads are flat').
Finally, we are still far from full 3D understanding of scenes. However, our work represents a step beyond 2.5D prediction and towards full 3D.

%\vspace{2mm}
\paragraph{\emph{\textbf{Acknowledgments.}}}
We would like to thank Tinghui Zhou and John Flynn for helpful discussions and comments. This work was done while ST was an intern at Google.

{\small
\bibliographystyle{splncs04}
\bibliography{references}
}

\clearpage
\section*{Appendix}
\section*{A1. Additional Visualizations}
\textbf{Dis-occlusions on KITTI.} We visualize in \figref{kitti_disocc} several sample dis-occlusion masks obtained for the KITTI dataset by running an occlusion-aware stereo algorithm. These dis-occlusion masks are only used to evaluate our learned LDI prediction, as we report the view synthesis error for estimated dis-occluded pixels (among other metrics).
%%%%%%%%%%%%%%%%%%%%%%%%%%%%%%%%%%%%%%%%%%
%%%%%%%%%%%%%%%%%%%%%%%%%%%%%%%%%%%%%%%%%%
\begin{figure*}
\centering
\includegraphics[width=1\textwidth]{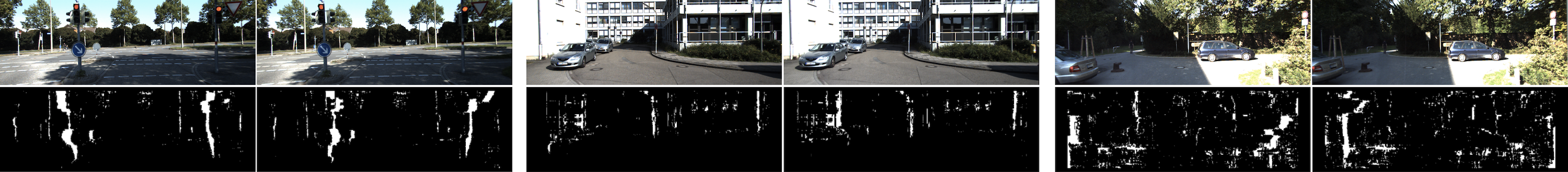}
\caption{\small \textbf{Dis-occlusion masks used for KITTI evaluation.} We use a stereo algorithm~\cite{Yamaguchi2014EfficientJS} to obtain pseudo ground-truth labels for dis-occlusion masks for images from the KITTI dataset. We visualize here the left and right stereo images (top) and the corresponding dis-occlusion masks. The dis-occlusions primarily represent regions corresponding to thin objects (poles in the first example), or object boundaries (\eg car in the second example), but are only approximate and can at times be erroneous (\eg the third example).}
\figlabel{kitti_disocc}
\end{figure*}

%%%%%%%%%%%%%%%%%%%%%%%%%%%%%%%%%%%%%%%%%%
%%%%%%%%%%%%%%%%%%%%%%%%%%%%%%%%%%%%%%%%%%

%\vspace{2mm}
%\noindent \textbf{Predictions for Full Images.} We visualized the LDI predictions for certain highlighted areas in the main text. We provide in \figref{synth_res} and \figref{kitti_res} additional visualizations, for the synthetic and KITTI dataset respectively, of the LDI predictions for full images.

%\begin{figure*}[t]
%\centering
%\includegraphics[width=1\textwidth]{figures/synth_results.png}
%\caption{\small \textbf{Sample LDI prediction results on synthetic data.} We show a,b) The ground-truth (top) and predicted (bottom) texture and inverse depth for the foreground layer, and c,d) The ground-truth (top) and predicted (bottom) texture for the second layer. Note that the input image is the same as the ground-truth texture image for the first layer.}
%\figlabel{synth_res}
%\end{figure*}

%%%%%%%%%%%%%%%%%%%%%%%%%%%%%%%%%%%%%%%%%%
%%%%%%%%%%%%%%%%%%%%%%%%%%%%%%%%%%%%%%%%%%

%\begin{figure*}[t]
%\centering
%\includegraphics[width=1\textwidth]{figures/kitti_results.png}
%\caption{\small \textbf{Sample LDI prediction results on the KITTI dataset.} We show a,b) The ground-truth (top) and predicted (bottom) texture and inverse depth for the foreground layer, and c,d) The ground-truth (top) and predicted (bottom) texture for the second layer. Note that the input image is the same as the ground-truth texture image for the first layer.}
%\figlabel{kitti_res}
%\end{figure*}

%%%%%%%%%%%%%%%%%%%%%%%%%%%%%%%%%%%%%%%%%%
%%%%%%%%%%%%%%%%%%%%%%%%%%%%%%%%%%%%%%%%%%

\section*{A2. Ablations}

We present ablations for some of the design decisions made for our training process. In particular, we demonstrate the importance of a) using the additional `min-view synthesis' loss for training, b) having a CNN architecture with disjoint prediction branches for predicting each LDI layer, and c) using a smoothness prior. We visualize in \figref{synth_ablations} the predictions made by CNNs trained without these. The `full' model represents the model described in the main text, `full - $L_{\mathrm{m-vs}}$' indicates the CNN trained without the additional  $L_{\mathrm{m-vs}}$ term in the training objective, and `full common' indicates a CNN without the disjoint prediction branches for each LDI layer (naturally, the final prediction weights are still different).

%%%%%%%%%%%%%%%%%%%%%%%%%%%%%%%%%%%%%%%%%%
%%%%%%%%%%%%%%%%%%%%%%%%%%%%%%%%%%%%%%%%%%
\renewcommand{\arraystretch}{1.2}
\setlength{\tabcolsep}{4pt}

\begin{table}
\centering
\begin{tabular}{l c c c c c}
\toprule
View Synthesis Error & Full  & Full $- L_{\mathrm{m-vs}}$ &  Full Common & Full $- L_{\mathrm{sm}}$\\
\midrule
All Pixels  & 0.0392  & 0.0411 & 0.0406 & 0.0414   \\
Dis-occluded Pixels & 0.1301  & 0.1414 & 0.1409 & 0.1352   \\
\bottomrule
\end{tabular}
\vspace{2mm}
\caption{\small \textbf{Ablations using view synthesis error on synthetic data}. The `full' model represents the model described in the main text, `full - $L_{\mathrm{m-vs}}$' indicates the CNN trained without the additional  $L_{\mathrm{m-vs}}$ term in the training objective, and `full common' indicates a CNN without the disjoint prediction branches for each LDI layer, and lastly, `full - $L_{\mathrm{sm}}$' indicates the CNN without a smoothness loss.}
\tablelabel{synthablate}
\vspace{-3mm}
\end{table}

%%%%%%%%%%%%%%%%%%%%%%%%%%%%%%%%%%%%%%%%%%
%%%%%%%%%%%%%%%%%%%%%%%%%%%%%%%%%%%%%%%%%%

\begin{figure*}
\centering
\includegraphics[width=\textwidth]{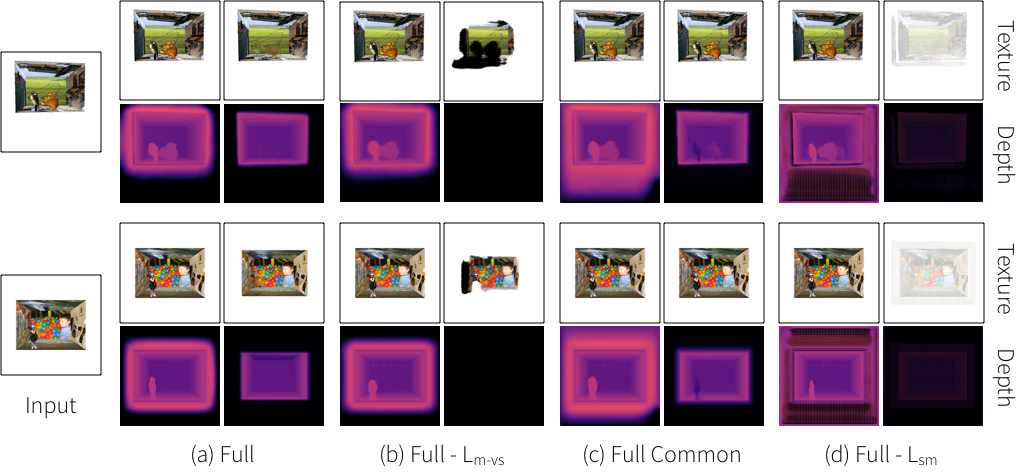}
\caption{\small \textbf{Inferred LDI representation on the synthetic data with different training variations.} We visualize some sample predictions on the synthetic data. The `full' model represents the model described in the main text, `full - $L_{\mathrm{m-vs}}$' indicates the CNN trained without the additional  $L_{\mathrm{m-vs}}$ term in the training objective, and `full common' indicates a CNN without the disjoint prediction branches for each LDI layer, and lastly, `full - $L_{\mathrm{sm}}$' indicates the CNN without a smoothness loss. See main text for more details.}
\figlabel{synth_ablations}
\end{figure*}

If we do not add the  $L_{\mathrm{m-vs}}$ term to the training objective, the learned CNN does not use the background layer at all---this occurs because there is initially very little learning signal for the background layer via the $L_{\mathrm{vs}}$ loss term as the (initially incorrect) foreground layer occludes it when projected to novel views. We also see that if the prediction branches are not disjoint, the background layer is not able to learn a meaningful amodal texture prediction, although it can learn to predict the (simpler) amodal geometry. We hypothesize that this occurs because the foreground  layer gets more learning signal, and sharing all the prediction weights makes it difficult for the leaning signals for the background layer to compete. Overall, we observe that both including the $L_{\mathrm{m-vs}}$ term, as well as having separate prediction branches for inferring each LDI layer, are important for learning a meaningful background layer prediction. Finally, the smoothness prior prevents undesirable artifacts.

\end{document}